\documentclass[runningheads]{llncs}
\usepackage{graphicx}
\usepackage{listings}
\usepackage{xcolor}
\usepackage{amsmath}
\usepackage{amsfonts}
\usepackage{bbm}
\usepackage{pifont}
\newcommand{\cmark}{\ding{51}}%
\newcommand{\xmark}{\ding{55}}%
\usepackage{multirow}
\usepackage{dsfont}
\usepackage{lipsum}
\usepackage[most]{tcolorbox}
\usepackage[colorlinks = true,
            linkcolor = blue,
            urlcolor  = blue,
            citecolor = blue,
            anchorcolor = blue]{hyperref}

\newtcolorbox{shadedcvbox}[1][]{enhanced jigsaw,
  colback=white!80!blue,
  coltext={black},
  boxrule=0pt,
  arc=3mm,
  auto outer arc,
  boxsep=3pt,
  left=4pt,
  right=2pt,
  bottom=2pt,
  top=2pt,
  fontupper={\bfseries},
  #1}

\definecolor{codegreen}{rgb}{0,0.6,0}
\definecolor{codegray}{rgb}{0.5,0.5,0.5}
\definecolor{codepurple}{rgb}{0.58,0,0.82}
\definecolor{backcolour}{rgb}{0.95,0.95,0.92}

\lstdefinestyle{mystyle}{
  backgroundcolor=\color{backcolour}, commentstyle=\color{codegreen},
  keywordstyle=\color{magenta},
  numberstyle=\tiny\color{codegray},
  stringstyle=\color{codepurple},
  basicstyle=\ttfamily\footnotesize,
  breakatwhitespace=false,         
  breaklines=true,                 
  captionpos=b,                    
  keepspaces=true,                 
  numbers=left,                    
  numbersep=5pt,                  
  showspaces=false,                
  showstringspaces=false,
  showtabs=false,                  
  tabsize=2
}

\title{Input Augmentation with SAM: 
Boosting Medical Image Segmentation with Segmentation Foundation Model}
\author{Yizhe Zhang\inst{1}, Tao Zhou\inst{1}, Shuo Wang\inst{2}, Peixian Liang\inst{3}, Danny Z. Chen\inst{3}}
\institute{School of Computer
Science and Engineering, Nanjing University of Science and Technology, Nanjing, Jiangsu 210094, China \\\email{yizhe.zhang.cs@gmail.com, taozhou.ai@gmail.com} \and Digital Medical Research Center, School of Basic Medical Sciences, Fudan University, Shanghai 200032, China \\\email{shuowang@fudan.edu.cn}\and Department of Computer Science and Engineering, University of Notre Dame, Notre Dame, IN 46556, USA \\\email{pliang@nd.edu, dchen@nd.edu}}
\titlerunning{SAMAug for Medical Image Segmentation}
\begin{document}
\maketitle
\begin{abstract}
The Segment Anything Model (SAM) is a recently developed large model for general-purpose segmentation for computer vision tasks. SAM was trained using 11 million images with over 1 billion masks and can produce segmentation results for a wide range of objects in natural scene images. SAM can be viewed as a general perception model for segmentation (partitioning images into semantically meaningful regions). Thus, how to utilize such a large foundation model for medical image segmentation is an emerging research target. This paper shows that although SAM does not immediately give high-quality segmentation for medical image data, its generated masks, features, and stability scores are useful for building and training better medical image segmentation models. In particular, we demonstrate how to use SAM to augment image input for commonly-used medical image segmentation models (e.g., U-Net). Experiments on three segmentation tasks show the effectiveness of our proposed SAMAug method. The code is available at \url{https://github.com/yizhezhang2000/SAMAug}.

\end{abstract}

\section{Introduction}

The Segment Anything Model (SAM) 
\cite{kirillov2023segment} is a remarkable recent advance in foundation models for computer vision tasks. SAM was trained using 11 million images and over 1 billion masks. Despite its strong capability in producing segmentation for a wide variety of objects, several studies~\cite{ji2023sam,deng2023segment,zhou2023can} showed that SAM is not powerful enough for segmentation tasks that require domain expert knowledge (e.g., medical image segmentation). 

\begin{figure}[h!]
\centering
\includegraphics[width=0.85\textwidth]{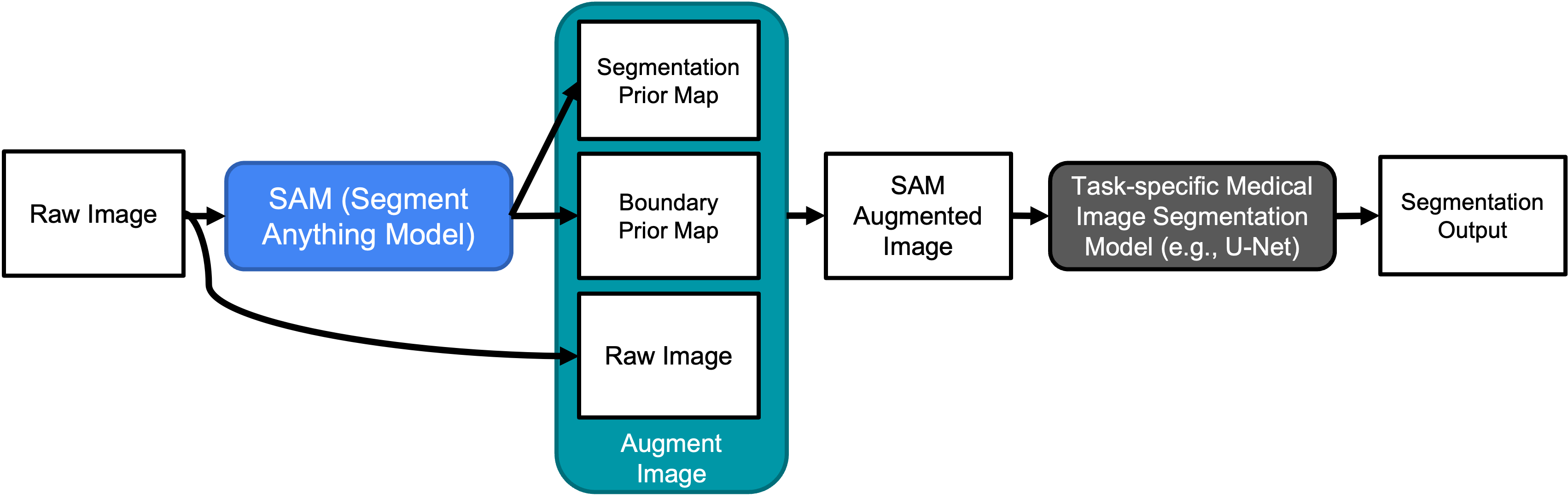}
\caption{Input augmentation with SAM for boosting medical image segmentation.} \label{fig:workflow}
\end{figure}

For a given medical image segmentation task with image and annotation pairs, we aim to build and train a medical image segmentation model, denoted by $\mathcal{M}$, on top of the segmentation foundation model SAM. We propose a new method called SAMAug that directly utilizes the segmentation masks (with stability scores) generated by SAM 
to augment the raw inputs of the medical image segmentation model $\mathcal{M}$. The input augmentation is performed by a 
fusion function. The inference process (with SAMAug) for a given image is illustrated in Fig.~\ref{fig:workflow}. The task-specific medical image segmentation model $\mathcal{M}$ is trainable using a specific dataset\footnote{SAMAug performs on all images, including training images and testing images.} (e.g., MoNuSeg~\cite{kumar2017dataset}). The parameters of SAM remain fixed, the fusion (augmentation) function is a parameter-free module, and the learning process aims to update the parameters of $\mathcal{M}$ with respect to the given foundation model SAM, the fusion function, and the training data. 

\textcolor{black}{Our main contributions can be summarized as follows. (1) We identify that the emerging segmentation foundation model SAM can provide attention (prior) maps for downstream segmentation tasks. (2) With a simple and novel method (SAMAug), we combine segmentation outputs of SAM with raw image inputs, generating SAM-augmented input images for building downstream medical image segmentation models. (3) We conduct comprehensive experiments to demonstrate that our proposed method is effective for both CNN and Transformer segmentation models in three medical image segmentation tasks.}

\section{Related Work}
\textbf{Data Augmentation.}
\textcolor{black}{Data augmentation (DA) has been widely used in training medical image segmentation models \cite{zhao2019data,chlap2021review}. A main aim of DA is to synthesize new views of existing samples in training data. Our SAMAug can be viewed as a type of DA technique. Unlike previous DA methods which often use hand-designed transformations (e.g., rotation, cropping), SAMAug utilizes a segmentation foundation model to augment raw images, aiming to impose semantically useful structures to the input of a medical image segmentation model. }

\noindent \textbf{Image Enhancement.}
\textcolor{black}{From the image enhancement (IE) view point, SAMAug enhances images by adding semantic structures from a segmentation foundation model. A critical difference between SAMAug and the previous enhancement methods~\cite{rundo2019medga,dinh2022new} is that traditional IE often works at a low level, e.g., de-blurring and noise reduction, and the purpose of enhancement is to reconstruct and recover. In contrast, SAMAug aims to add high-level structures to raw images, providing better semantics for the subsequent medical image segmentation model.}

\noindent \textbf{Recent SAM-related Methods.}
\textcolor{black}{Since the introduction of SAM, many attempts have been made to understand and utilize SAM for medical image analysis (e.g., \cite{huang2023segment,zhou2023can,ma2023segment,wu2023medical,zhang2023segment,zhang2023survey,zhang2023understanding,qiao2023robustness,gao2023desam,zhang2023comprehensive}). Recent work has shown that SAM alone, without further fine-tuning and/or adaptation, often delivers unsatisfied results for medical image segmentation tasks \cite{huang2023segment,zhou2023can}. In order to utilize SAM more effectively, Ma et al.~\cite{ma2023segment} proposed to fine-tune SAM using labeled images. Wu et al.~\cite{wu2023medical} proposed to add additional layers to adapt SAM for a medical image segmentation task. Compared with these fine-tuning and adaptation methods, our method is more efficient in computation and memory costs during model training. In test time, these fine-tuning, adapting, and augmentation methods all require performing forward propagation of test images through SAM.}

\section{Methodology}
In Section~\ref{sec:seg_and_boundary_PM}, we describe the two key image representations obtained by applying SAM to a medical image, a segmentation prior map and a boundary prior map. In Section~\ref{sec:AII}, we show how to augment a medical image using the two obtained prior maps. In Section~\ref{sec:MT}, we present the details of using augmented images in training a medical image segmentation model. Finally, in Section~\ref{sec:modeltest}, we show how to use the trained model in model deployment (model testing).

\begin{figure}[t]
\centering
\includegraphics[width=1.0\textwidth]{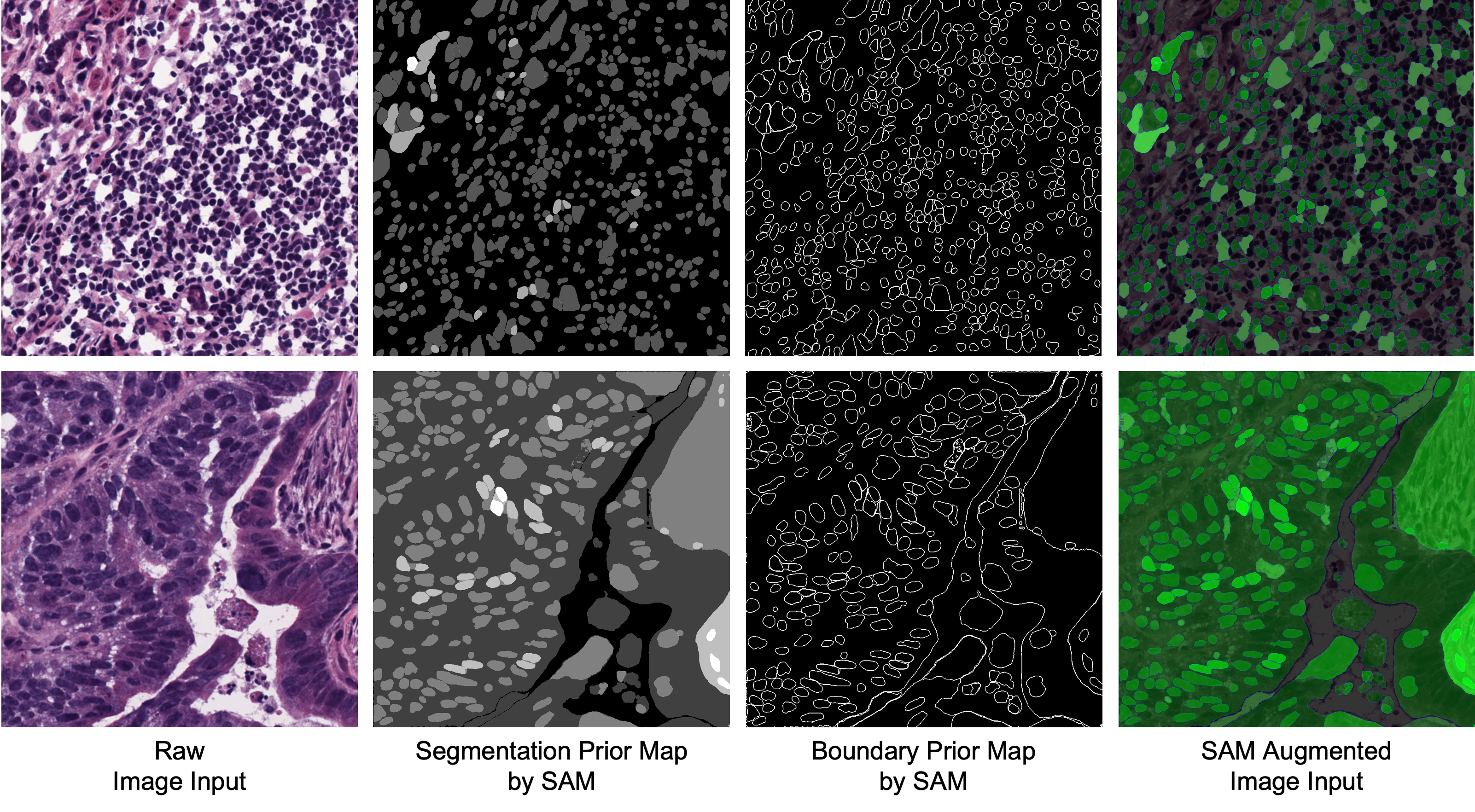}
\caption{Visual examples of a raw input image, its segmentation prior map by SAM, boundary prior map by SAM, and SAM-augmented image input (illustrated in Fig.~\ref{fig:workflow}). The image sample is from the MonuSeg dataset~\cite{kumar2017dataset}.} \label{fig:samaug_vis_samples}
\end{figure}

\subsection{Segmentation and Boundary Prior Maps}\label{sec:seg_and_boundary_PM}
In the grid prompt setting, SAM uses a grid prompt to generate segmentation masks for a given image. That is, segmentation masks are generated at all plausible locations in the image. The generated segmentation masks are then stored in a list. For each segmentation mask in the list, we draw the mask on a newly created segmentation prior map using the value suggested by the mask's corresponding stability score (generated by SAM). In addition to the segmentation prior map, we further generate a boundary prior map according to the masks provided by SAM. We draw the exterior boundary of each segmentation mask in the mask list and put all the boundaries together to form a boundary prior map. For a given image $x$, we generate two prior maps, $\textrm{prior}_{\textrm{seg}}$ and $\textrm{prior}_{\textrm{boundary}}$, using the process discussed above. In Fig.~\ref{fig:samaug_vis_samples} (the second and third columns), we give visual examples of these two prior maps thus generated.

\subsection{Augmenting Input Images}\label{sec:AII}
With the prior maps generated, our next step is to augment the input image $x$ with the generated prior maps. We choose a simple method for this augmentation: adding the prior maps to the raw image. Note that many medical image segmentation tasks can be reduced to a three-class segmentation task in which the 1st class corresponds to the background, the 2nd class corresponds to the regions of interest (ROIs), and the 3rd class corresponds to the boundaries between the ROIs and background. We add the segmentation prior map to the second channel of the raw image and the boundary prior map to the third channel of the raw image. If the raw image is in gray-scale, we create a 3-channel image with the first channel consisting of the gray-scale raw image, the second channel consisting of its segmentation prior map (only), and the third channel consisting of its boundary prior map (only). For each image $x$ in the training set, we generate its augmented version $x^{aug}=\textrm{Aug}(\textrm{prior}_\textrm{seg},\textrm{prior}_\textrm{boundary},x)$.  Fig.~\ref{fig:samaug_vis_samples} (the fourth column) gives a visual example of the SAM-augmented image input.

\subsection{Model Training with SAM-Augmented Images}\label{sec:MT}
\textcolor{black}{With the input augmentation on each image sample in the training set, we obtain a new augmented training set $\{(x^{aug}_1,y_1), (x^{aug}_2,y_2), \dots, (x^{aug}_n,y_n)\}$, where $x^{aug}_i\in \mathbb{R}^{w\times h \times 3}$, $y_i \in \{0,1\}^{w\times h \times C}$ is the annotation of the input image $x_i$, and $C$ is the number of classes for the segmentation task. A common medical image segmentation model $\mathcal{M}$ (e.g., a U-Net) can be directly utilized for learning from the augmented training set. A simple way to learn from SAM-augmented images is to use the following learning objective with respect to the parameters of $\mathcal{M}$:}
\begin{equation}
\sum_{i=1}^{n} loss (\mathcal{M}(x^{aug}_i), y_i) .
\label{eq-loss-simple}
\end{equation}

\textcolor{black}{The above objective only uses SAM-augmented images for model training. Consequently, in model testing, the trained model accepts only images augmented by SAM. In situations where SAM fails to give plausible prior maps, we consider training a segmentation model using both raw images and images with SAM augmentation. The new learning objective is to minimize the following objective with respect to the parameters of $\mathcal{M}$:}
\begin{equation}
\sum_{i=1}^{n} \beta  loss (\mathcal{M}(x_i), y_i) + \lambda loss (\mathcal{M}(x^{aug}_i), y_i),
\label{eq-loss}
\end{equation}
\textcolor{black}{where $\beta$ and $\lambda$ control the importance of the training loss for samples with raw images and samples with augmented images. When setting $\beta$ = 0 and $\lambda$ = 1, the objective function in Eq.~(\ref{eq-loss}) is reduced to Eq.~(\ref{eq-loss-simple}). By default, we set both $\beta$ and $\lambda$ equal to 1. The spatial cross-entropy loss or Dice loss can be used for constructing the loss function in Eq.~(\ref{eq-loss-simple}) and Eq.~(\ref{eq-loss}). An SGD-based optimizer (e.g., Adam~\cite{kingma2014adam}) can be applied to reduce the values of the loss function.}

\subsection{Model Deployment with SAM-Augmented Images}\label{sec:modeltest}
\textcolor{black}{When the segmentation model is trained using only SAM-augmented images, the model deployment (testing) requires the input also to be SAM-augmented images. The model deployment can be written as:}
\begin{equation}
\hat{y}=\tau(\mathcal{M}(x^{aug})),
\label{eq:tt_samaugonly}
\end{equation}
\textcolor{black}{where $\tau$ is an output activation function (e.g., a sigmoid function, a softmax function), and $x^{aug}$ is a SAM-augmented image (as described in Section~\ref{sec:AII}). When the segmentation model $\mathcal{M}$ is trained using both raw images and SAM-augmented images, we identify new opportunities in inference time to fully realize the potential of the trained model. A simple way of using $\mathcal{M}$ would be to apply sample inference twice for each test sample: The first time inference uses the raw image $x$ as input and the second time inference uses its SAM augmented image as input. The final segmentation output can be generated by the average ensemble of the two outputs. Formally, this inference process can be written as:}
\begin{equation}
\hat{y}=\tau(\mathcal{M}(x)+ \mathcal{M}(x^{aug})).
\label{eq:ttsimple}
\end{equation}

\textcolor{black}{Another way of utilizing the two output candidates $\mathcal{M}(x)$ and $\mathcal{M}(x^{aug})$ is to select a plausible segmentation output from these two candidates:}
\begin{equation}
\hat{y}=\tau(\mathcal{M}(x^*)),
\label{eq:ttselect}
\end{equation}
\textcolor{black}{where $x^*$ is obtained via solving the following optimization:}
\begin{equation}
x^*={\rm argmin}_{x'\in\{x, x^{aug}\}}Entropy(\tau(\mathcal{M}(x'))).
\label{eq:ttselect}
\end{equation}
\textcolor{black}{Namely, we choose an input version out of the two input candidates ($x$ and $x^{aug}$) according to the entropy (prediction certainty) of the segmentation output. Segmentation output with a lower entropy means that the model is more certain in its prediction, and a higher certainty in prediction often positively correlates to higher segmentation accuracy~\cite{wangtent}.}

\section{Experiments and Results}
\subsection{Datasets and Setups}
We perform experiments on the Polyp~\cite{zhou2023can}, MoNuSeg~\cite{kumar2017dataset}, and GlaS~\cite{sirinukunwattana2017gland} benchmarks to demonstrate the effectiveness of our proposed SAMAug method. 
For the polyp segmentation experiments, we follow the training setup used in training the state-of-the-art (SOTA) model HSNet~\cite{zhang2022hsnet}\footnote{\url{https://github.com/baiboat/HSNet}}. For the MoNuSeg and GlaS segmentation, the training of a medical image segmentation model uses the Adam optimizer~\cite{kingma2014adam}, with batch size = 8, image cropping window size = 256 $\times$ 256, and learning rate = $5e-4$. The total number of training iterations is 50K. The spatial cross entropy loss is used for the model training.

\begin{figure}[h!]
\centering
\includegraphics[width=0.8\textwidth]{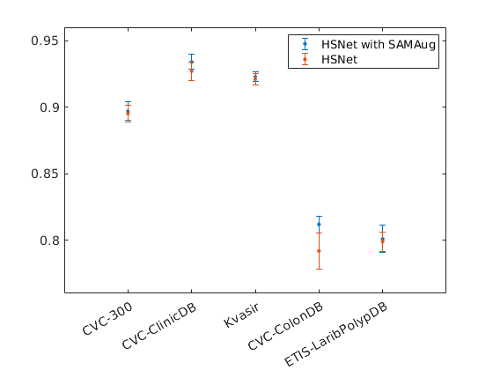}
\caption{Polyp segmentation results of the vanilla HSNet and SAMAug-enhanced HSNet.} \label{fig:Polpyseg_HSNET}
\end{figure}

\subsection{Polyp Segmentation on Five Datasets}
\begin{figure}[h!]
\centering
\includegraphics[width=0.8\textwidth]{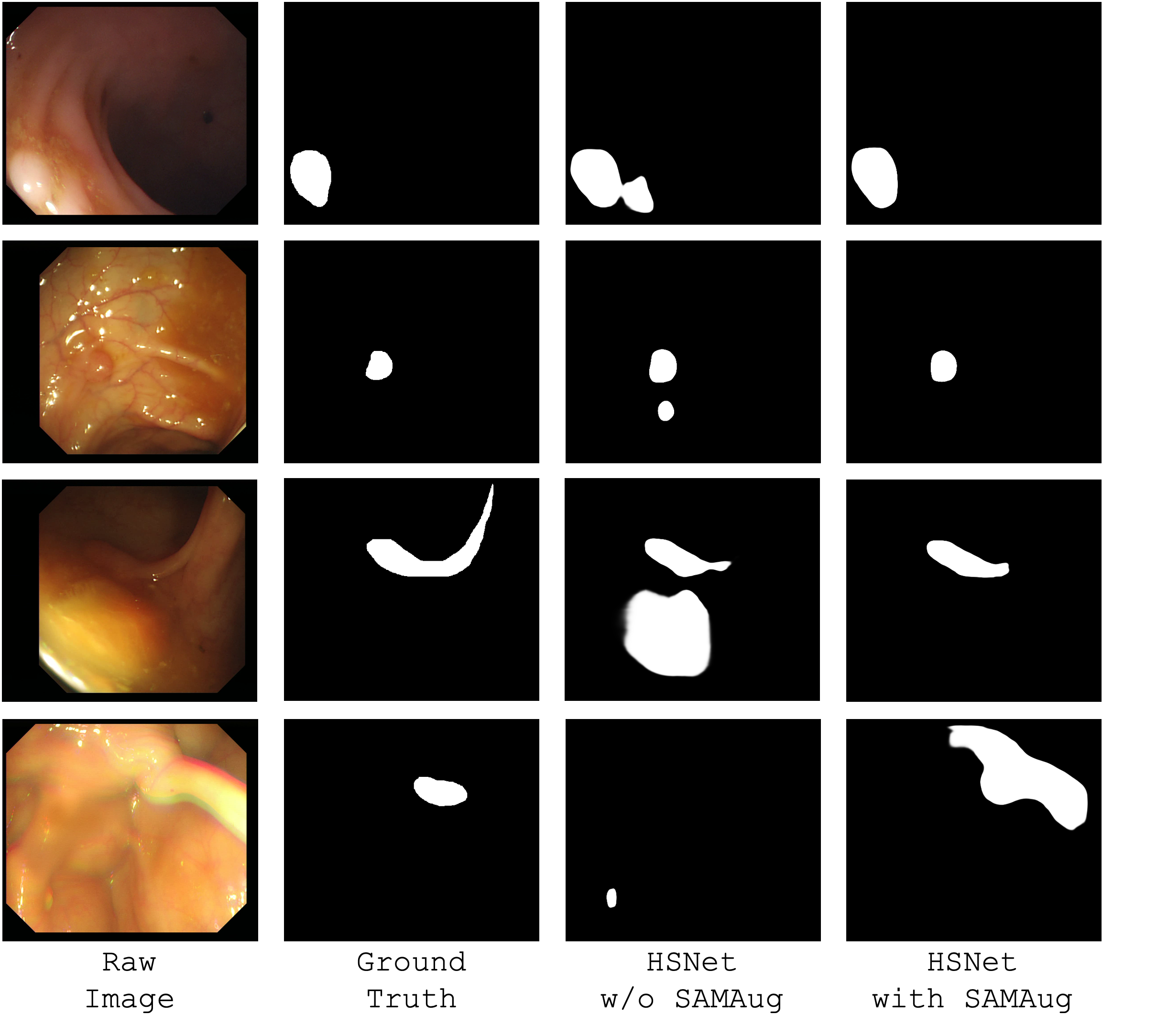}
\caption{Visual result comparisons of the vanilla HSNet and SAMAug-enhanced HSNet in polyp segmentation.} \label{fig:Polpyseg_HSNET_vis}
\end{figure}

\textcolor{black}{Automatic polyp segmentation in endoscopic images can help improve the efficiency and accuracy in clinical screenings and tests for gastrointestinal diseases. Many deep learning (DL) based models have been proposed for robust and automatic segmentation of polyps. Here, we utilize the SOTA model HSNet~\cite{zhang2022hsnet} for evaluating our proposed SAMAug method. We use the objective function described in Eq.~(\ref{eq-loss}) in model training. In test time, we use the model deployment strategy given in Eq.~(\ref{eq:ttselect}). In Fig.~\ref{fig:Polpyseg_HSNET}, we show the segmentation performance (in Dice score) of the vanilla HSNet and SAMAug-enhanced HSNet on the test sets of CVC-300~\cite{vazquez2017benchmark}, CVC-ClinicDB~\cite{bernal2015wm}, Kvasir~\cite{jha2020kvasir}, CVC-ColonDB~\cite{tajbakhsh2015automated}, and ETIS~\cite{silva2014toward}. All the model training sessions were run ten times with different random seeds for reporting the means and standard deviations of the segmentation performance. In Fig.~\ref{fig:Polpyseg_HSNET}, we observe that SAMAug improves HSNet on the CVC-ClinicDB and CVC-ColonDB datasets significantly, and remains at the same level of performance on the other three datasets (all validated by t-test). Furthermore, we give visual result comparisons in Fig.~\ref{fig:Polpyseg_HSNET_vis}. }

\subsection{Cell Segmentation on the MoNuSeg Dataset}
\begin{figure}[h!]
\centering
\includegraphics[width=1.0\textwidth]{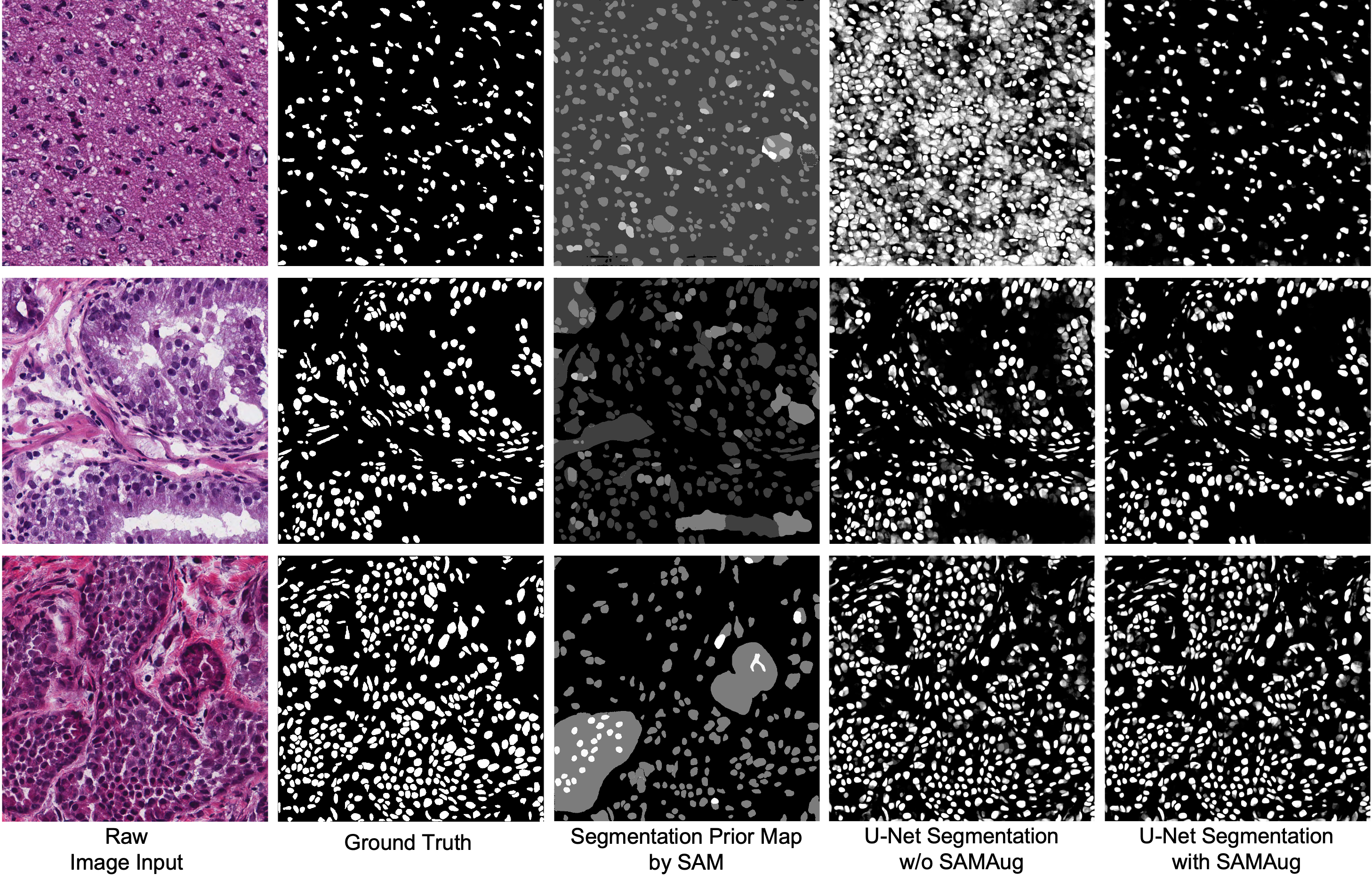}
\caption{Visual comparisons of segmentation results on the MoNuSeg dataset.} \label{fig:cellseg_vis_results}
\end{figure}

The MoNuSeg dataset~\cite{kumar2017dataset} was constructed using H\&E stained tissue images (at 40x magnification) from the TCGA archive~\cite{wang2016practical}. The training set consists of 30 images with about 22000 cell nuclear annotations. The test set contains 14 images with about 7000 cell nuclear annotations. We use the objective function described in Eq.~(\ref{eq-loss-simple}) in model training. In test time, we use the model deployment strategy given in Eq.~(\ref{eq:tt_samaugonly}). In Table~\ref{cell_results}, we show clear advantages of our proposed method in improving segmentation results for the U-Net, P-Net, and Attention U-Net models. AJI (Aggregated Jaccard Index) is a standard segmentation evaluation metric\footnote{\url{https://monuseg.grand-challenge.org/Evaluation/}} used on MoNuSeg which evaluates segmentation performance on the object level. F-score evaluates the cell segmentation performance on the pixel level. In addition, we give visual result comparisons in Fig.~\ref{fig:cellseg_vis_results}. Note that, although the segmentation generated by SAM (e.g., see the 3rd column of Fig.~\ref{fig:cellseg_vis_results}) does not immediately give accurate cell segmentation, SAM provides a general segmentation perceptual prior for the subsequent DL models to generate much more accurate task-specific segmentation results.
\begin{table}[t]
\centering
\caption{Cell segmentation results on the MoNuSeg dataset.}
\begin{tabular}{|c|c|c| c|}
\hline
Model & SAMAug & AJI & F-score \\\hline
{Swin-UNet~\cite{cao2023swin}} & \xmark & 61.66  & 80.57 \\
\hline
\multirow{2}{*}{U-Net~\cite{ronneberger2015u}} & \xmark & 58.36 & 75.70\\
\cline{2-4}
 & \cmark & 64.30 & 82.36\\
\hline
 \multirow{2}{*}{P-Net~\cite{wang2018deepigeos}} & \xmark & 59.46 & 77.09\\
\cline{2-4}
 & \cmark & 63.98 & 82.56\\
\hline
 \multirow{2}{*}{Attention UNet~\cite{oktay2018attention}} & \xmark & 58.76 & 75.43\\
\cline{2-4}
 & \cmark & 63.15 & 81.49\\
\hline
\end{tabular}
\label{cell_results}
\end{table}


\subsection{Gland Segmentation on the GlaS Dataset}
The GlaS dataset~\cite{sirinukunwattana2017gland} has 85 training images (37 benign
(BN), 48 malignant (MT)), and 60 test images (33 BN, 27 MT) in part A and 20
test images (4 BN, 16 MT) in part B. We use the official evaluation code\footnote{\url{https://warwick.ac.uk/fac/cross_fac/tia/data/glascontest/evaluation/}} for evaluating segmentation performance. For simplicity, we merge test set part A and test set part B, and perform segmentation evaluation at once for all the samples in the test set. We use the objective function described in Eq.~(\ref{eq-loss-simple}) in model training. In test time, we use the model deployment strategy given in Eq.~(\ref{eq:tt_samaugonly}). From Table~\ref{gland_results}, one can see that U-Net with SAMAug augmentation performs considerably better than that without SAMAug augmentation. 

\begin{table}[t]
\centering
\caption{Gland segmentation results on the GlaS dataset.}
\begin{tabular}{|c|c|c|c|}
\hline
Model & SAMAug & F-score & Object Dice \\\hline
\multirow{2}{*}{U-Net~\cite{ronneberger2015u}} & \xmark & 79.33 & 86.35\\
\cline{2-4}
 & \cmark & 82.50 & 87.44\\
\hline
\end{tabular}
\label{gland_results}
\end{table}

\section{Conclusions}
\textcolor{black}{In this paper, we proposed a new method, SAMAug, for boosting medical image segmentation that uses the Segment Anything Model (SAM) to augment image input for commonly-used medical image segmentation models. Experiments on three segmentation tasks showed the effectiveness of our proposed method. Future work may consider conducting further research on: (1) designing a more robust and advanced augmentation function; (2) improving the efficiency of applying SAM in the SAMAug scheme; (3) utilizing SAMAug for uncertainty estimations and in other clinically-oriented applications.}

\bibliographystyle{plain}
\bibliography{sample}

\end{document}